\documentclass{article}

\ifdefined\pdfinfoomitdate
\fi
\ifdefined\pdfsuppressptexinfo
\fi

\usepackage[preprint]{neurips_2026}

\usepackage[utf8]{inputenc}
\usepackage[T1]{fontenc}

\usepackage{microtype}
\usepackage{amsmath,amssymb,amsfonts}
\usepackage{booktabs}
\usepackage{tabularx}
\usepackage{array}
\newcolumntype{Y}{>{\raggedright\arraybackslash}X}
\newcolumntype{L}[1]{>{\raggedright\arraybackslash}p{#1}}
\usepackage{enumitem}
\usepackage{xcolor}
\usepackage{graphicx}
\usepackage{flafter}
\usepackage{float}
\usepackage{placeins}
\usepackage{url}
\usepackage{hyperref}
\usepackage[nameinlink,capitalize,noabbrev]{cleveref}

\hypersetup{
  pdftitle={Prune to Probe: Causal Decision-Stability Probes for Efficient Reasoning},
  pdfauthor={Yufei Qi and Chonghe Jiang},
  colorlinks=true,
  linkcolor=blue!55!black,
  citecolor=blue!55!black,
  urlcolor=blue!55!black
}

\newcommand{\method}{\textsc{Prune-to-Probe}}
\newcommand{\basepolicy}{\textsc{BehavioralExit}}
\newcommand{\shampolicy}{\textsc{RandomProbe}}
\newcommand{\rcc}{\textsc{P2P-MC}}

\newcommand{\artifacturl}{\url{https://github.com/YQ-HA/prune_to_probe}}

\title{Prune to Probe:\\
Causal Decision-Stability Probes for Efficient Reasoning}

\author{%
  Yufei Qi \\
  New York University \\
  \texttt{yq2179@nyu.edu}
  \And
  Chonghe Jiang \\
  Massachusetts Institute of Technology \\
  \texttt{chonghej@mit.edu}
}

\begin{document}
\raggedbottom
\maketitle
\begin{abstract}
Reasoning models often continue generating after producing an answer, yet stopping prematurely may interrupt useful verification or revision. In this work, we investigate whether targeted interventions in a model's internal computation can provide a reliable signal for when to stop reasoning. We introduce \method{}, an inference-time controller that uses causal activation interventions to probe the stability of a model's current answer. Specifically, we identify late-MLP coordinates that contribute to post-answer continuation while excluding those important for answer completion. Once an answer is stated, \method{} temporarily suppresses the selected coordinates and measures the resulting stopping response relative to matched random interventions. Crucially, these interventions serve only as probes: the model either returns its existing answer or resumes generation from an untouched cache. On MATH, \method{} reduces reasoning-token forwards by 2.60$\%$, preserving all naturally correct trajectories while repairing six errors. On MMLU-Pro, it achieves 1.145$\%$ pooled savings with no observed harm. Further analysis shows that the intervention captures decision stability rather than correctness (AUROC 0.502). These results demonstrate the potential of causal activation probes for efficient inference-time control without permanent model modification.
\end{abstract}

\section{Introduction}
Reasoning models often continue generating after producing an answer. While some of this additional reasoning may be redundant, it can also serve to verify, revise, or improve the current answer. This creates a fundamental challenge for early-exit methods: how can we determine when to stop reasoning without observing future tokens or knowing the correct answer? Existing behavioral signals alone may not fully capture whether the model has settled on its current decision. We therefore investigate whether the model's internal computation can provide additional evidence for early-exit decisions.

Our key insight is that the causal response to an intervention can be more useful for inference-time control than the intervention itself. Through matched ablations, we identify late-MLP coordinates that influence answer completion and post-answer continuation. However, these effects are state-dependent: coordinates that contribute to continued reasoning at one point may support useful computation later. Consequently, permanently suppressing these coordinates can disrupt subsequent generation. Rather than using pruning to modify the model, we use it to probe the model's computation at a particular inference state.

Building on this insight, we introduce \method{}, an early-exit controller based on bounded causal interventions. Once the model has stated an answer, we temporarily suppress selected continuation-related coordinates while excluding those important for answer completion. We then compare the resulting stopping response against layer- and count-matched random interventions to assess whether the current state exhibits a mechanism-specific tendency to stop. Crucially, all intervened branches are discarded after probing. The controller either returns the existing answer or resumes generation from the untouched cache, leaving the model's natural generation process unchanged.

We evaluate \method{} on mathematical reasoning and multiple-choice tasks. On held-out MATH, it reduces fully charged reasoning-token forwards by 2.60$\%$, preserving all naturally correct trajectories while repairing six errors. Across two prospective MMLU-Pro evaluations, a frozen multiple-choice variant achieves 1.145$\%$ pooled savings with no observed harm to naturally correct trajectories. Further analysis reveals an important distinction: the intervention response reflects decision stability rather than answer correctness (AUROC 0.502). These findings demonstrate that localized causal interventions can provide useful signals for inference-time control, even when the identified coordinates cannot be safely removed throughout generation.

\section{Related work}

Weight pruning and reasoning-aware compression preserve model quality under resource
constraints \citep{frantar2023sparsegpt,sun2023wanda,meng2024alps,
lucas2025rac,xiang2025ssgr}. Mechanistic methods analyze activations and use activation
or edge interventions to identify causal components
\citep{conmy2023acdc,syed2024eap,hanna2024faithfulness}. Methods for efficient reasoning
control response length, steer activations, or learn when to stop generation
\citep{aggarwal2025l1,azizi2026asc,liu2026neat,sun2026refrain}. Our controller differs
from behavior-only exit rules because it tests a causally selected target against
matched-random interventions at the current state. The intervention response is the
computational trace used for the exit decision. Unlike checkpoint compression, the
method does not change the stored model or the free-running path.

\section{Method}

Our method uses causal activation interventions to assess whether a
reasoning model can stop after producing an answer. The method consists of
two stages. First, we identify the internal coordinates that causally
influence the completion of an answer and the continuation after an answer is produced. Second,
we temporarily suppress selected coordinates and evaluate the model's
response to assess decision stability. It is important to note that these interventions only serve as diagnostic probes and do not modify the model's
subsequent generation.

We measure the stopping response using the next-token logits
$\ell$. For Qwen models, the stop token is the single-token
\texttt{</think>} boundary. We define the stop margin as
\[
    h = \ell_{\mathrm{stop}}
    - \max_{v \ne \mathrm{stop}} \ell_v,
\]
where $h \geq 0$ indicates that the stop token has the highest
next-token logit. This definition is used throughout our experiments.

\paragraph{Causal localization.}
We first identify late-MLP coordinates associated with two
computational functions: answer completion and post-answer
continuation. Using matched ablations, we classify a coordinate
as \emph{completion-positive} if removing it increases the
negative log-likelihood of answer tokens, and as
\emph{continuation-positive} if its removal increases the
stop margin after a correct answer has been given. We use leave-one-prompt-out ranking over the final eight
MLP blocks for coordinate selection, with layer- and count-matched random ablations
serving as controls.

Our intervention target consists of continuation-positive
coordinates that are not completion-positive. This selection
aims to isolate computation associated with continued reasoning
while minimizing interference with answer completion.
These categories describe measured causal effects rather than
distinct semantic circuits, and their roles may change across
generation states.

For Qwen3-8B, we zero 32 \texttt{down\_proj} input coordinates
in each of layers 28--35, yielding 256 coordinates in total.
For Qwen3-1.7B, we select 16 coordinates per layer across
layers 20--27, yielding 128 coordinates. All interventions
are matched by layer and coordinate count.

\paragraph{Decision-stability probe.}
Once the model has stated a completed answer, we assess its
current state through a bounded intervention. We temporarily
suppress the selected continuation-related coordinates and
compare the resulting stop margin against two matched-random
interventions.

Let $h_B$ denote the original stop margin, $h_T$ the margin
under the targeted intervention, and $h_{R_1}, h_{R_2}$
the margins under the two random interventions. We define
\begin{equation}
    c = h_T - \max(h_{R_1}, h_{R_2}),
    \qquad
    d = \max_{i \in \{1,2\}} |h_{R_i} - h_B|.
    \label{eq:main-certificate}
\end{equation}
Here, $c$ measures the intervention-specific increase in
the tendency to stop, relative to random suppressions.
The quantity $d$ measures sensitivity to the random
interventions, allowing us to reject states in which
generic perturbations substantially affect the stop margin.

Together, these quantities provide a causal, state-dependent
signal for early-exit decisions. The intervention is applied
only during probing, and all intervened branches are discarded
afterward. If the controller accepts the current answer, it
returns the existing answer; otherwise, generation resumes
from the untouched cache.

\paragraph{MATH controller.}
For mathematical reasoning, we first apply a reference-free
answer-structure check. When a prompt explicitly requests
multiple values, the controller rejects a scalar singleton
as a potentially incomplete answer.

For candidates that pass this check, the controller exits
directly if the original stop margin satisfies $h_B \geq 0$.
Otherwise, it applies the decision-stability probe and
accepts an early exit when
\[
    c \geq 6,
    \qquad
    d \leq 0.75.
\]
If neither condition is satisfied, generation continues
from the original cache.

We evaluate the controller against two matched baselines.
The structure-aware baseline applies the same answer-structure
check, but without the targeted intervention. The equal-cost
random baseline replaces the target with a third, matched-random
intervention. All probe forwards, including those for rejected
states, are included in compute accounting.

\paragraph{Multiple-choice controller.}
For multiple-choice reasoning, we introduce a more conservative
variant, \rcc{}, that combines causal probing with behavioral
consistency checks. Before probing, the model must have stated
the same answer option at least twice, generated at least 64
additional tokens, and exhibited no unresolved revision or
oscillation. The trigger must be at or before the frozen
position bound of 3231 tokens, and the original model must still
favor continuation ($h_B<0$).

Eligible states undergo the same decision-stability probe,
but with stricter calibration requirements:
\[
    c \geq 0,
    \qquad
    d \leq 0.4375.
\]
States passing these conditions receive two additional
five-token forced-choice readouts, one from the original
model and one under the targeted intervention.

Let $m_B(a)$ and $m_T(a)$ denote the candidate-to-best-alternative
logit margins for the current answer $a$ under the original
and intervened models, respectively. We require $a$ to remain
the top-ranked option in both readouts, with
\[
    m_B(a) \geq 6,
    \qquad
    m_B(a) - m_T(a) \geq 1.
\]
These conditions require both behavioral consistency and
a measurable response to the targeted intervention.
The three stop probes cost 3 token-forwards. States rejected
at this stage incur no forced-choice readout cost. If both
five-token readouts are executed, the total cost is 13
token-forwards, whether the state is accepted or rejected.

As in the MATH controller, all interventions are discarded
after probing, and free-running generation always follows
the original model. Full controller definitions are provided
in \cref{app:full-method}.

\begin{figure}[ht]
\centering
\fbox{\begin{minipage}{0.94\textwidth}
\small
\textbf{Bounded MATH controller.}
Given a completed candidate and its untouched prefix cache:
\begin{enumerate}[leftmargin=*,itemsep=1pt,topsep=2pt]
  \item If the candidate violates the prompt-derived answer-structure contract,
  continue natural generation without probing.
  \item Compute $h_B$. If $h_B\geq0$, copy the visible candidate and stop.
  \item Otherwise compute the target and two calibration-random margins.
  \item If $c\geq6$ and $d\leq0.75$, copy the candidate; otherwise resume from the
  untouched cache.
\end{enumerate}
Every executed one-token probe is charged. No intervened branch becomes the generation
cache.
\end{minipage}}
\caption{The copy-on-write decision rule used for the held-out MATH test.}
\label{fig:method}
\end{figure}

\FloatBarrier
\section{Experimental design}

We evaluate \method{} on mathematical reasoning and multiple-choice
tasks using two models from the Qwen3 family.

For mathematical reasoning, we use Qwen3-8B \citep{yang2025qwen3}
in BF16 with a fixed CoT-Wanda pruning mask \citep{sun2023wanda}.
The mask is constructed using layer-sequential Wanda on 32
trajectories from the GSM8K training split, each truncated to
4096 tokens. Row-wise 30\% unstructured pruning yields a realized
global sparsity of 29.9847\%. We use GSM8K for causal coordinate
discovery and evaluate the resulting controller on disjoint MATH
prompts \citep{cobbe2021gsm8k,hendrycks2021math}. No validation
or held-out prompts are used during mask construction. To ensure
an unbiased evaluation, we freeze the complete controller before
held-out testing and evaluate it on 256 held-out prompts only
after all predefined criteria are satisfied on 128 validation
prompts. Both evaluations use decoding seeds 42 and 43.

To assess transfer beyond mathematical reasoning, we evaluate
dense Qwen3-1.7B on MMLU-Pro \citep{wang2024mmlupro}, excluding
its mathematics category. We freeze the multiple-choice controller
before conducting two prospective evaluations, each comprising
1,024 questions with no overlap between their prompt sets.
The second evaluation is conducted only after the first satisfies
all predefined criteria, without modifying the controller.

We document the complete experimental sequence, including
unsuccessful development experiments, in \cref{app:design}.
For reproducibility, we provide a public repository
containing derived intervention records, reference code, and
scripts for recomputing record-based controller outcomes and
selected reported statistics.\footnote{\artifacturl}

\section{Results}

\begin{table}[H]
\centering
\small
\setlength{\tabcolsep}{3.5pt}
\begin{tabularx}{\textwidth}{@{}L{2.2cm}Yrr@{}}
\toprule
Evaluation & Probe exits (correct / wrong) & Harm / repair & Charged savings \\
\midrule
8B MATH held-out & 86 (86 / 0) & 0 / 6 & 99.55 (2.60\%) \\
1.7B MMLU-Pro validation & 99 (64 / 35) & 0 / 4 & 35.58 (1.267\%) \\
1.7B MMLU-Pro confirmation & 91 (55 / 36) & 0 / 1 & 31.12 (1.032\%) \\
\bottomrule
\end{tabularx}
\caption{Main results. Savings are fully charged reasoning-token forwards per trajectory;
percentages use natural reasoning forwards as the denominator. A harm
changes a naturally correct trajectory to a wrong policy output; a repair changes a
naturally wrong trajectory to a correct policy output.}
\label{tab:main-results}
\end{table}

\begin{figure}[H]
  \centering
  \includegraphics[width=\textwidth]{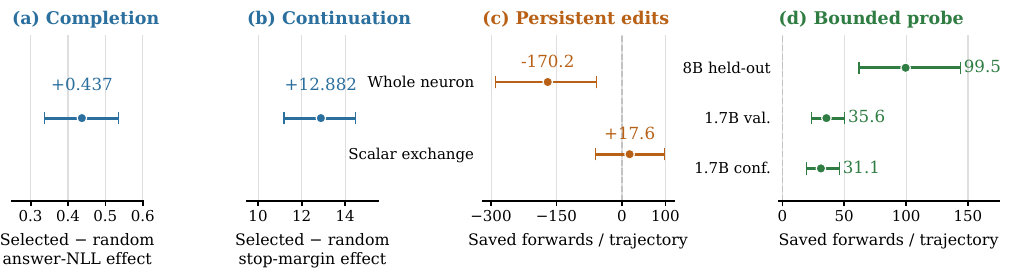}
  \caption{The experimental progression. Panels (a--b) show selected-minus-random local
  effects in their original units. Panel (c) shows that persistent edits do not produce
  reliable savings. Panel (d) reports fully charged savings from the bounded probe.
  Points are means and bars are 95\% bootstrap intervals; positive savings indicate
  lower inference cost.}
  \label{fig:evidence-summary}
\end{figure}

\paragraph{Local effects do not justify persistent pruning.}
Matched ablations confirm that the identified coordinates
causally influence both answer completion and post-answer continuation
(\cref{fig:evidence-summary}). However, these effects are local
to the states in which they are measured and do not necessarily
persist throughout generation. Protecting entire neurons
increases reasoning length, whereas scalar mask exchange yields
no statistically reliable savings and fails to outperform matched
random exchange. These findings suggest that causal relevance in one state does not imply the safe modification of the same coordinates throughout generation, which motivates our use of
bounded interventions as diagnostic probes.

\paragraph{Held-out MATH evaluation.}
We evaluate \method{} on 512 trajectories from 256 held-out
MATH prompts. The controller accepts 86 early exits, all of
which preserve correct answers. It retains all 428 naturally
correct trajectories while repairing six errors. After accounting
for all intervention overhead, \method{} saves an average of
99.55 reasoning-token forwards per trajectory, corresponding
to a 2.60\% reduction. The 95\% prompt-cluster bootstrap
interval is $[61.84, 143.57]$ forwards saved. The improvements
over both the structure-aware baseline and the equal-cost
random-intervention control are also statistically resolved.
An independent implementation reproduces all controller
decisions and compute accounting.

\paragraph{Non-math transfer and decision stability.}
We further evaluate \method{} on two prospective MMLU-Pro
populations using a frozen multiple-choice controller.
Across the 2,048 trajectories, the controller saves an
average of 33.35 reasoning-token forwards per trajectory,
or 1.145\%, with a 95\% interval of $[24.51, 43.37]$.
Both evaluations satisfy their predefined criteria, with
no observed degradation of naturally correct trajectories
and five natural errors repaired.

These results also reveal an important limitation of the
intervention signal. Among 190 accepted exits, 71 preserve
incorrect answers, most of which would also remain incorrect
under natural generation. The causal stop contrast achieves
a correctness AUROC of 0.502, indicating little ability to
distinguish correct from incorrect candidates. Thus, the
intervention provides evidence of decision stability rather
than answer correctness. Additional analyses of the stopping
and forced-choice responses are provided in the appendix.

\paragraph{Functional-subspace analysis.}
We additionally investigate whether the selected coordinates
support distinct forms of post-answer computation. A predefined
pilot examines 96 states categorized by prefix-visible evidence
of repetition, verification, revision, or alternative search.
Although the targeted intervention consistently affects the
stop margin across these categories, the differences between
categories do not replicate across decoding seeds. The
predefined feasibility criterion is therefore not met, and
we do not proceed to held-out evaluation. These findings
support an aggregate continuation-sensitive intervention
but do not establish distinct functional circuits for the
four proposed behaviors.

\FloatBarrier
\section{Discussion and limitations}

Our results demonstrate that localized causal interventions
can provide useful signals for inference-time control without
permanently altering a model. Instead of removing components
to accelerate generation, \method{} uses their causal
effects to characterize the current reasoning state. The
comparison against matched random interventions helps isolate
a mechanism-specific stopping response. The bounded
probe ensures that, when needed, subsequent generation proceeds from the untouched model state. This distinction
is essential because the computational roles of individual
coordinates can change throughout generation.

Nevertheless, the interpretation of these signals requires
care. The selected coordinates are continuation-sensitive
under the localization objective; however, this does not imply that they form
independent semantic circuits or that they are dispensable in other inference states. In particular, the intervention
target is distinct from the weights removed in constructing
the sparse checkpoint and does not represent the entire
complement of the completion-related coordinates. The target may also influence various forms of post-answer computation, such as repetition, verification, revision, and searching for alternatives.
Our functional-subspace pilot does not reliably separate these behaviors. Finally, the intervention response reflects the stability of the model's decision, not its correctness. Since a stable decision can be incorrect, the response should not be used as a standalone verifier of correctness.

Our evaluation is limited to two model sizes within the
Qwen3 family, including one sparse Qwen3-8B checkpoint,
MATH, and 13 non-mathematical MMLU-Pro domains. We do
not evaluate other architectures, languages, or open-ended
tasks. Although no naturally correct trajectories are harmed
in the final evaluations, this observation does not establish
a general safety guarantee. For the multiple-choice
evaluations, the one-sided 95\% upper bound on harm is
0.294\% per naturally correct source trajectory and 2.59\%
among accepted naturally correct exposures. Furthermore,
the reported 1.145\% savings measure fully charged
reasoning-token forwards rather than end-to-end latency.
The functional-subspace pilot is also limited by its small
sample size and behavior-based state categorization.
Further details and unsuccessful development protocols are
reported in \cref{app:full-results,app:limits}.

Taken together, these findings support causal intervention
responses as state-dependent signals for reasoning control.
Their value lies not in identifying permanently removable
components or verifying answer correctness, but in providing
additional evidence for when generation can be terminated.

\section{Responsible-use statement}

An early-exit system may preserve an incorrect answer or prevent the computation that would correct it. The acceptance rule should not be used to verify correctness, and deploying it in high-stakes settings would require task-specific validation, calibrated
abstention, monitoring, and a reliable fallback. These experiments use public reasoning
benchmarks and do not involve personal or sensitive data.

\section{Conclusion}

We introduce \method{}, an inference-time controller that uses
bounded causal interventions to evaluate the stability of decisions in reasoning
models. Rather than permanently suppressing continuation-related activations, our approach probes them, enabling early-exit decisions
while preserving the model's original generation path. Experiments
on MATH and MMLU-Pro demonstrate reductions in reasoning computation
without observed degradation of naturally correct trajectories in
the final evaluations. Our findings underscore the potential of
causal activation probes for inference-time control. Internal computations can provide valuable insights into when to cease reasoning,
even if they do not indicate answer correctness or facilitate permanent model modifications.

\bibliographystyle{plainnat}
\bibliography{references}

\appendix

\section{Additional related work}

\paragraph{Weight pruning and reasoning-aware compression.}
SparseGPT, Wanda, ALPS, GPTQ, and AWQ preserve model quality under sparsity or
precision constraints \citep{frantar2023sparsegpt,sun2023wanda,meng2024alps,
frantar2023gptq,lin2024awq}. Reasoning-aware calibration and component selection further
account for on-policy reasoning distributions
\citep{lucas2025rac,xiang2025ssgr,sheth2026cap,choi2026laquant}. Our final method does
not claim a smaller stored checkpoint: it temporarily zeros selected activations for a
bounded counterfactual and leaves generation on the original route.

\paragraph{Mechanistic circuit discovery and pruning.}
Mechanistic interpretability methods identify causal components through activation or
edge interventions \citep{conmy2023acdc,syed2024eap,hanna2024faithfulness}. Related
pruning methods exploit contextual sparsity, preserve identified functional networks,
or allocate sparsity across layers
\citep{liu2023dejavu,liu2025functionalnetworks,chen2025dlp}. We use mechanistic ranking
to define an intervention and matched negative controls. A selected coordinate is not
claimed to be a complete semantic circuit; the causal claim comes from the controlled
intervention effect.

\paragraph{Efficient reasoning and early exit.}
Length control, activation steering, learned exit signals, and optimal-stopping methods
can shorten reasoning \citep{aggarwal2025l1,azizi2026asc,liu2026neat,
sun2026refrain,ehab2026ospruner,lan2026crisp}. Our focus is narrower: whether a
mechanism-ranked pruning intervention supplies decision evidence beyond an otherwise
equivalent behavioral exit. The controller therefore includes both a non-pruning
baseline and an equal-cost random-intervention sham.

\section{Full method and controller definitions}
\label{app:full-method}

\subsection{Intervention scope and causal localization}

The intervention acts on a specific set of MLP intermediate coordinates. It does not
refer to all weights removed from the sparse checkpoint, and it does not include every
coordinate that the selection procedure leaves unselected. We call a coordinate
\emph{completion-positive} when its ablation increases answer-token negative
log-likelihood on completion examples. We call it \emph{continuation-positive} when its
ablation reduces the best-nonstop-minus-stop margin after a correct answer commitment.
The online target contains only continuation-positive coordinates that are not
completion-positive. We refer to this set as the completion-safe continuation target.

Completion examples are correct trajectories at an answer-token prediction event.
Continuation examples contain a correct answer commitment followed by at least 512
additional reasoning tokens. Discovery employs leave-one-prompt-out selection for the final
eight MLP \texttt{down\_proj} inputs. Each causal test compares the selected coordinates
with random sets that are matched by layer and count.

These labels describe effects under the discovery objectives and do not establish two
disjoint semantic modules. A selected coordinate may contribute to different operations
at another position or on another task. For example, the continuation target may facilitate mechanical repetition in one state and useful verification or revision in another. Our
experiments measure the aggregate local effect of the selected set without decomposing
these possible functions.

\subsection{From persistent pruning to a bounded probe}

Initially, we tried editing the sparse mask while leaving an activation intervention active
during generation. Both approaches extend the intervention beyond the states used for
localization. These approaches may suppress coordinates after their function changes and redirect subsequent sampling with small
logit changes. Functional heterogeneity within the
target set is another possible explanation for the failures: if overlapping coordinates
support low-value repetition, verification, revision, and alternative searches, then a
persistent intervention can suppress all of these operations together.

We therefore restrict the final intervention to the current state. The probe tests whether the stop margin depends on the completion-safe continuation target and then discards
the intervened branch. If an exit is not supported, generation resumes from the
untouched cache. The controller does not assume that every targeted coordinate is
permanently dispensable.

At the earliest completed explicit answer assertion, let $h_B$ be the unperturbed
stop-versus-best-nonstop margin defined in the main text, $h_T$ the margin under the target continuation intervention,
and $h_{R_1},h_{R_2}$ the margins under two disjoint matched-random interventions. Define
\begin{align}
  c &= h_T - \max(h_{R_1},h_{R_2}), \\
  d &= \max\{|h_{R_1}-h_B|, |h_{R_2}-h_B|\}.
\end{align}
The null-calibrated rule accepts only if
\begin{equation}
  c \geq 6.0 \quad\text{and}\quad d \leq 0.75.
  \label{eq:certificate}
\end{equation}
The first condition requires a large mechanism-specific response. The second requires
the two negative-control interventions to behave as local nulls at that state. A third
disjoint random intervention replaces the target intervention in an equal-cost sham.

\subsection{Enumeration-aware answer contract}

Equation~\ref{eq:certificate} addresses the causal stopping response, rather than whether the
candidate satisfies the requested output cardinality. Before any intervention, \method{} verifies an observable answer-interface contract. Direct plural enumeration requests
require a top-level list, set, prompt-requested interval, union, plus/minus form,
empty-set/no-solutions form, or textual disjunction. Aggregates such as the sum, product,
or number of all values remain scalar-eligible. A direct plural request followed by a
scalar singleton abstains conservatively.

The final cascade is shown in \cref{fig:cascade-rule}.
\begin{figure}[t]
\centering
\fbox{\begin{minipage}{0.93\columnwidth}
\small
\textbf{Enumeration-aware null-calibrated readiness}
\begin{enumerate}[leftmargin=*,itemsep=1pt,topsep=3pt]
  \item Given completed candidate $a$, prefix state $s$, and prompt $x$: if $a$
  violates the observable answer-structure contract for $x$, continue the untouched
  natural route.
  \item Compute the unperturbed stop margin $h_B$. If $h_B\geq0$, copy $a$ without
  intervention probes.
  \item Otherwise compute the target and two calibration-random intervention margins.
  \item If $c\geq6.0$ and $d\leq0.75$, copy $a$; otherwise continue the untouched
  natural route.
\end{enumerate}
\end{minipage}}
\caption{The exact online cascade.}
\label{fig:cascade-rule}
\end{figure}

Each executed target or random intervention is charged as one token-forward equivalent.
Rejected states continue from the untouched base cache. Accepted states copy the
already completed candidate rather than sampling from the intervened route.

\subsection{Multiple-choice causal coupling}

Multiple-choice transfer requires a different interface contract. \rcc{} first abstains at the earliest
explicit choice convergence followed by a 64-token dwell if the
observed prefix contains an unresolved answer revision or post-commitment oscillation, or
if the trigger lies beyond position 3231, which is no longer supported by discovery. It also requires at
least two prefix-visible assertions of the same candidate $a$. These checks use neither
the reference answer nor future tokens.

For an otherwise eligible state, let $h_B,h_T,h_{R_1},h_{R_2}$ denote the base, target,
and two matched-random stop margins. The base must still favor continuation
($h_B<0$). The stop margins must also satisfy
\begin{equation}
 h_T-\max(h_{R_1},h_{R_2})\geq0,
 \qquad
 \max_i |h_{R_i}-h_B|\leq0.4375.
 \label{eq:rcc-stop}
\end{equation}
Passing states receive two five-token forced-answer readouts using the same dense
prefix cache, one unperturbed and one with the target coordinates zeroed. Let $m_B(a)$
and $m_T(a)$ be the candidate logit minus the strongest alternative-choice logit. Both
readouts must rank $a$ first, and
\begin{equation}
 m_B(a)\geq6,\qquad m_B(a)-m_T(a)\geq1.
 \label{eq:rcc-answer}
\end{equation}
We treat the answer as invariant only when the target intervention measurably reduces
its margin. The target is used for the probe and the forced-answer suffix, but not for
free-running generation. The three stop probes cost 3 token-forwards.
States rejected by the stop-margin gate incur only this cost; states that reach
the two five-token readouts incur 10 additional forwards, for a total of 13.
All executed probe and readout forwards are charged, including those on rejected states.

\subsection{Controls and accounting}

We compare \method{} with the following: (i) natural inference; (ii) \basepolicy{}, which uses the
same answer contract and accepts only when $h_B\geq0$; and (iii) \shampolicy{}, which
uses the same cascade and cost, but substitutes a third random intervention for the target.
The causal-specificity statistic compares the target margin with that of the strongest of all the three random interventions. Correctness transitions are paired with each trajectory's frozen
natural outcome. A \emph{harm} is natural-correct to policy-wrong, and a \emph{repair} is
natural-wrong to policy-correct.

The primary computation metric is the number of fully charged reasoning-token forwards. Every executed
probe is included, even on rejected states. We report paired bootstrap intervals within
a decoding seed and prompt-cluster bootstrap intervals when multiple seeds share a
prompt. Generated forward savings are not necessarily wall-clock savings; therefore, a
separate hardware replay diagnostic measures the exact online branch primitives.

\section{Full experimental design}
\label{app:design}

\paragraph{Model and tasks.}
The primary checkpoint is Qwen3-8B \citep{yang2025qwen3} in BF16 with a fixed
29.9847\% unstructured CoT-Wanda mask. Layer-sequential Wanda uses 32 GSM8K training
trajectories of at most 4096 tokens; the nominal row-wise sparsity is 30\%.
Mechanistic discovery uses GSM8K
\citep{cobbe2021gsm8k}; the ambiguous-answer controller is evaluated on disjoint MATH
prompts \citep{hendrycks2021math}. Non-mathematical transfer uses 13 domains of
MMLU-Pro \citep{wang2024mmlupro}, excluding its math category. The generation cap is
8192 reasoning tokens.

\paragraph{Evaluation sequence.}
We do not promote development populations to confirmation, and we keep failed fresh
tests. After fixing the enumeration-aware MATH rule, a joint split reserves 128
prompts for validation and 256 disjoint prompts for held-out evaluation. Each prompt is
run with seeds 42 and 43, yielding 256 and 512 trajectories. The held-out population is
evaluated once, after every frozen validation gate passes. The answer contract, selected
coordinates, thresholds, intervention dose, controls, accounting, and gates remain unchanged.

Additional MATH experiments only change the decoding seeds, replay measured branch
primitives on an L40S, or compare matched dense and sparse endpoints. These are
post-confirmation diagnostics. The smaller-model experiments rediscover coordinates for
Qwen3-1.7B. Tests of the initial controller and a subsequent position guard reveal harms.
We then recalibrate the random-null cap on those opened populations and test
the resulting controller unchanged on 256 new prompts.

The non-MATH study retains all failed development and repair attempts. Its final
multiple-choice rule is fixed before two category-stratified populations are generated.
Each population contains 1,024 questions, uses a different decoding seed, and is disjoint
from the other population and from all development data. The second population is
evaluated only after the first passes every scientific gate, with no method change.

\paragraph{Final gates.}
The final protocol requires zero observed natural-correct harms, a minimum number of
correct pruning accepts, a pruning-accept error rate below unconditional first-assertion
exit, correctness no worse than both controls, resolved fully charged savings against
natural, \basepolicy{}, and \shampolicy{}, and positive target specificity. All gates
must pass separately by seed and under the prompt-cluster analysis.

\section{Additional results}
\label{app:full-results}

\subsection{Local causality does not imply persistent control}

Removing the selected completion coordinates raises answer-token NLL by 0.4371 more than
matched random ablation, with 95\% interval $[0.3376,0.5343]$. Removing selected
continuation coordinates lowers the continuation margin by 12.8815 more than random,
interval $[11.1916,14.4642]$. Both intervals exclude zero and establish a local effect
under the corresponding discovery objective.

The same coordinates do not provide reliable control when the edits persist.
Whole-neuron protection adds 170.16 tokens per prompt for 128 fresh prompts, with an interval of $[59.01,289.27]$. A Pareto-safe scalar mask exchange saves 172.80 tokens on 128 held-out prompts during the initial evaluation. However, a replication with 512 prompts yields an average savings of only 17.62 tokens, with an interval of $[-60.57,97.95]$ in savings orientation, and
does not outperform a random exchange from an eligible pool. A regime audit reveals large savings on
baseline-long trajectories and a resolved penalty on the ordinary complement, which
cancel each other out. Persistent activation interventions also redirect trajectories sampled later. These results demonstrate a local effect, but they do not show that the selected
coordinates are dispensable throughout the generation process. Their function may change across
states, or the target may contain several types of post-answer computation. The bounded
probe does not depend on either explanation.

\subsection{Null calibration is necessary but not sufficient}

The original null-calibrated rule passes a predefined validation test and a
single-seed held-out test. The held-out test has 59 pruning accepts, 58 of which were correct. There were zero
natural-correct harms, three repairs, and 146.72 charged forward savings (3.79\%), with an interval of $[93.65,207.77]$ in savings orientation. The rule saves the same 146.72 forwards relative to
the non-pruning base-stop control and 149.34 relative to the equal-cost random sham. The
target intervention exceeds the strongest random intervention by 4.1547 logits, interval
$[3.7781,4.5260]$.

The unchanged seed-43 replay remains efficient and specific, but it harms one natural correct trajectory by accepting an incomplete singleton for an all-solutions request.
A narrow comma-list guard then fails on a fresh all-values prompt. We retain both
failures as negative results because the intervention response does not determine whether
a candidate exhausts the requested answer set. We develop the enumeration-aware contract
from these failures and fix it before opening a new validation population.

\subsection{Final Qwen3-8B confirmation}

The two-seed validation of the complete enumeration-aware cascade includes 58 correct pruning
accepts and five repairs, with no natural-correct harms. Every efficiency,
control, and specificity gate passes. We then open the predefined held-out test without
changing the method.

The held-out evaluation contains 256 prompts and 512 trajectories. All 86
probe-supported exits are correct. The controller preserves all 428 naturally correct
trajectories and repairs six natural errors. Fully charged compute falls by 99.55
forwards per trajectory (2.60\%), with
prompt-cluster interval $[61.84,143.57]$ in savings orientation. Savings versus
\basepolicy{} are also 99.55, interval $[61.97,144.02]$, and savings versus
\shampolicy{} are 101.72, interval $[64.04,146.29]$. Specificity is 4.6722 logits,
interval $[4.4551,4.8891]$. An independent implementation reconstructs every
decision, charged total, correctness transition, source hash, interval, and gate.

The answer contract abstains from probing on three validation states and seven
held-out states. These held-out abstentions include correct singleton candidates that
could have saved compute, as well as incomplete candidates that should not be accepted.
This conservative interface rule sacrifices the associated savings.

\subsection{Post-confirmation boundary evidence}

Changing only the decoding seeds to 44/45 yields two additional passing runs. Across
the four-seed descriptive union of 1,024 trajectories, 185/186 pruning acceptances are correct. There are zero natural-correct harms and ten repairs. Mean fully charged
savings are 104.63 forwards with a prompt-cluster interval of $[73.54,140.65]$. These runs provide additional decoding-seed evidence on the same prompts; they do not constitute another held-out population.

A hardware diagnostic times the exact online single-request KV-cache path on an L40S
and projects the median cost of primitives over 512 trajectories. The hardware-calibrated projection
saves 10.34 seconds (7.03\%) per trajectory, with a savings interval of $[8.38,12.41]$ seconds. Serializing all detector work still saves 7.03 seconds (4.78\%), with a savings interval of $[4.95,9.25]$. Four direct branch replays agree with the projected direction.
The projection excludes batching, queueing, networking, streaming, and production serving.

The endpoint diagnostic applies the unchanged, sparse-discovered coordinates to 128
matched, historical prompt--seed pairs at dense and 29.9847\%-sparse endpoints. The sparse arm has 26/26
correct pruning accepts, zero harms, one repair, and 149.09-forward savings (3.81\%).
The dense arm has 28/28 correct accepts, zero harms, two repairs, and 227.64-forward
savings (5.32\%). Both arms resolve all controls and specificity tests. The interval for
the dense-minus-sparse difference crosses zero; therefore, the experiment supports transfer
across the two tested endpoints without establishing dense superiority or a full
sparsity curve.

\subsection{Qwen3-1.7B: mechanism transfer and a sharper controller boundary}

The smaller model study rediscovers coordinates from scratch in dense Qwen3-1.7B rather
than copying the 8B indices. On 28 completion examples, the selected-minus-random
answer-NLL effect is 0.8979, with an interval of $[0.7438,1.0579]$. On 15 continuation examples, the margin effect is 16.9909, with an interval of $[14.0664,19.8775]$. Both causal gates and an independent tensor/record audit
pass.

The initial 128-trajectory MATH diagnostic saves 782.33 charged forwards (18.04\%), with an interval of $[574.41,1010.03]$ in savings orientation, and improves correctness from
103/128 to 108/128 through six repairs and one harm. Specificity is 7.4081, interval
$[6.8764,7.9012]$. However, only 60/63 pruning acceptances are correct. The resulting 4.76\% accepted-error rate exceeds the 3.54\% error rate of unconditional first-assertion exit. The zero-harm and error-rate
gates fail. The experiment provides partial cross-model evidence but does not confirm
the controller on a second model.

\begin{table}[t]
\centering
\small
\setlength{\tabcolsep}{3pt}
\begin{tabular}{@{}lrr@{}}
\toprule
 & Qwen3-8B held-out & Qwen3-1.7B initial test \\
\midrule
Evaluation trajectories & 512 & 128 \\
Correct pruning accepts & 86/86 & 60/63 \\
Harms / repairs & 0 / 6 & 1 / 6 \\
Charged saving & 2.60\% & 18.04\% \\
Specificity & 4.6722 & 7.4081 \\
Strict protocol & pass & fail \\
\bottomrule
\end{tabular}
\caption{Results at two model scales. The percentages are not a head-to-head comparison:
the models have different natural lengths, neuron doses, and evaluation status.}
\label{tab:cross-model}
\end{table}

After opening the initial controller diagnostic, we observe that the sole harm occurs
later than every commitment position in the 15-example continuation-discovery support.
A candidate guard that disables the intervention branch beyond the discovery maximum of
3231 would remove the observed harm while retaining 41/42 correct pruning accepts, two
repairs, and zero harms.
It saves 541.54 charged forwards per trajectory (12.49\%), with interval
$[357.59,754.81]$ in savings orientation; supported-probe specificity is 7.3464, interval
$[6.6735,7.9587]$. The independently audited analysis is post hoc. It specifies a
model-scaled hypothesis for a future test and does not repair the failed initial
protocol.

A fresh prompt-disjoint population of 128 trajectories tests the unchanged guard under
seed 43. The guard fails. The controller saves 529.97
charged forwards (12.07\%), interval $[355.19,729.39]$ in savings orientation, and its
specificity remains 6.7824, but only 35/39 pruning accepts are correct. Correctness moves
from 102/128 to 101/128 through two repairs and three natural-correct harms. All harms
occur before position 3231. The independently audited result shows that position support
is not a sufficient safety condition in this population, even though every efficiency
and specificity gate passes.

The three harmful acceptances have large target-versus-random contrasts
(8.4219--9.3594 logits), yet their matched random deviations are relatively small (0.4688--0.5625). Thus, a large intervention-specific readiness response can occur when the initial response is incorrect. A predeclared descending-grid analysis selects a stricter
model-specific random-null cap of 0.4375 on the two development populations. At that cap, the initial population has 30/30 correct pruning accepts, zero harms, two repairs, and 390.37 forward savings. The new position-guard population has 26/26 correct pruning accepts, zero harms, one repair, and
352.19 forward savings. The calibration remains post hoc and defines the next frozen
1.7B hypothesis, but it does not establish second-model confirmation.

A final test freezes that exact 0.4375 cap, the 3231 support bound, all selected
coordinates, gates, and seed 44 before generating a new 256-prompt population. All 54
probe-supported exits are correct. No naturally correct trajectory is harmed, and two
natural errors are repaired, raising correctness from 205/256 to 207/256. The target saves 346.94 fully
charged forwards per trajectory (8.48\%), interval $[244.17,460.87]$ in savings
orientation, with resolved advantages over natural, the structure-aware base, and the
equal-cost sham. Specificity is 7.1304, interval $[6.6692,7.5699]$. Every frozen gate
and all 19 independent provenance/recomputation checks pass. This fresh population
confirms the calibrated controller. The mechanism and efficiency signal transfer, but
position support alone is insufficient for safety. This model requires a calibrated
random-null cap.

\subsection{Non-math transfer: causal stability does not identify truth}

The MMLU-Pro experiments test whether behavioral convergence and the response to the
target intervention distinguish correct from incorrect answers. The frozen behavioral-stability rule passes one development population but then fails its predefined replication. The
replication accepts 16 correct candidates and 15 wrong candidates, with one harm and one repair.
The model can converge on a stable incorrect answer. Requiring the candidate identities to survive both base and target readouts provides little filtering: this occurs in 96.5\% of the first population's eligible states and in all of the replication's eligible states, although none of the jointly tested threshold rules pass. The rule in
\cref{eq:rcc-answer} therefore also requires a nontrivial reduction in candidate margin.

Two prospective populations test the resulting \rcc{} rule. They use dense Qwen3-1.7B
in BF16, the full 8192-token cap, different decoding seeds, and disjoint
category-stratified questions. All guards, selected coordinates, intervention doses,
thresholds, accounting rules, and seven gates are identical. We evaluate the
confirmation population only after the validation population passes its scientific
gates and independent audit.

\begin{table*}[t]
\centering
\small
\setlength{\tabcolsep}{4pt}
\begin{tabularx}{\textwidth}{@{}Yrrr@{}}
\toprule
Metric & Validation & Confirmation & Descriptive union \\
\midrule
Trajectories & 1,024 & 1,024 & 2,048 \\
Natural correct & 538 & 481 & 1,019 \\
Accepted: correct / wrong & 64 / 35 & 55 / 36 & 119 / 71 \\
Natural-correct harms / repairs & 0 / 4 & 0 / 1 & 0 / 5 \\
Charged saving per trajectory & 35.58 & 31.12 & 33.35 \\
Relative charged saving & 1.267\% & 1.032\% & 1.145\% \\
95\% interval, savings orientation & $[23.62,50.09]$ & $[19.19,45.82]$ & $[24.51,43.37]$ \\
Criteria / recomputation checks & 7/7; 23/23 & 7/7; 23/23 & --- \\
\bottomrule
\end{tabularx}
\caption{Prospective non-math validation and conditional confirmation. The union is a
descriptive analysis conducted after confirmation, not a third confirmatory test. Probe and
forced-readout forwards are fully charged.}
\label{tab:nonmath-results}
\end{table*}

Both prospective runs pass the frozen gates. In the descriptive union, policy
correctness moves from 1,019/2,048 to 1,024/2,048 through five repairs and no observed
harms (one-sided exact paired improvement $p=0.03125$). The saving is 68,296
token-forwards, or 33.35 per trajectory, with interval $[24.51,43.37]$ in savings
orientation. The one-sided 95\% upper bound on harm is 0.294\% per naturally correct
source trajectory and 2.59\% conditional on the 114 accepted naturally correct
exposures.
All 13 categories have positive point-estimate savings and zero observed harms, but this
does not imply homogeneous reliability: accepted-error point estimates range from
12.5\% (1/8; engineering) to 75.0\% (3/4; history). The small category-level acceptance
counts make these subgroup extremes descriptive rather than powered safety estimates.

Across 414 intervention-probed states, the target stop response exceeds the stronger
matched-random response by 7.928 logits on average, interval $[7.549,8.313]$. Across 231
forced-answer readouts, the same intervention reduces the candidate margin by 1.979 logits,
interval $[1.828,2.130]$, and accepted states preserve candidate identity. The stop
causal contrast has correctness AUROC 0.502, which is close to chance. Base and target
answer margins have AUROCs of 0.665 and 0.668. Accepted error is 9.07 percentage points
below raw convergence, interval $[-15.20,-2.90]$, but only 1.76 points below the
post-guard comparator, with an unresolved interval $[-7.48,3.81]$.

The intervention response identifies a reproducible decision-stability state but does
not determine whether the candidate is correct. The controller combines this signal
with prefix guards and a fallback to the natural route. A post-confirmation threshold
surface shows the cost of
demanding a larger answer-margin response. Raising the required answer-margin drop from
1 to 2 logits lowers accepted error from
37.37\% to 30.56\%, but reduces charged saving from 1.145\% to 0.600\%. This is
hypothesis-generating only and does not redefine the confirmed rule.

\subsection{Functional-subspace heterogeneity pilot}

The predefined pilot reuses two MMLU-Pro development populations generated under
seeds 54 and 55. Prefix-only rules assign mutually exclusive repetition, verification,
explicit revision, and alternative
states, with priority revision $>$ alternative $>$ verification $>$ repetition. A fixed,
outcome-blind hash order selects 12 prompts per state and source revision, giving 96
unique prompts. At each prefix, one dense cache is shared across the unperturbed
base, the unchanged 128-coordinate Qwen3-1.7B target, and three layer/count-matched
random interventions.
Each branch records the immediate stop and forced-choice margins and a bounded 64-token
diagnostic continuation. The natural generation path and controller are unchanged.

\begin{table}[t]
\centering
\small
\setlength{\tabcolsep}{3pt}
\begin{tabularx}{\textwidth}{@{}L{1.55cm}rrY@{}}
\toprule
State & Seed 54 stop effect & Seed 55 stop effect & Candidate-margin effect, seed 54 / 55 \\
\midrule
Repetition & $10.269$ & $7.826$ & $-1.563$ / $-2.260$ \\
Verification & $8.864$ & $7.376$ & $-2.233$ / $-2.264$ \\
Revision & $7.185$ & $8.657$ & $-0.694$ / $-0.934$ \\
Alternative & $8.501$ & $6.566$ & $-1.840$ / $-2.236$ \\
\bottomrule
\end{tabularx}
\caption{Target minus the mean of three matched-random interventions in the functional-state
pilot. Each cell contains 12 prompts. Full bootstrap intervals are reported in the
supplementary statistics.}
\label{tab:functional-pilot}
\end{table}

The primary repetition-versus-pooled stop-margin contrast is $+2.086$ under seed 54,
with bootstrap interval $[-0.132,4.283]$ and Holm-adjusted $p=.2734$. It is $+0.293$
under seed 55, interval $[-2.750,3.341]$ and adjusted $p=.8514$. Stop-margin omnibus
tests are also negative. Immediate candidate-margin heterogeneity passes correction
under seed 54 ($p=.00270$), where revision states have a smaller drop, but does not
replicate under seed 55 ($p=.29324$). Immediate top choices never change under the target
intervention (0/96), and only one target top choice changes after the bounded branch.
Lexical branch events are
too sparse for a functional claim. Independent recomputation reproduces the statistics,
but the predefined feasibility criterion fails; no held-out experiment is run.

\section{Additional mechanistic discussion}

The discovery procedure selects a completion-safe and continuation-positive target. This
target is neither the full set of weights removed from the sparse checkpoint nor the
complement of the protected completion set. Local interventions demonstrate that the target affects
continuation at the localization states. However, they do not prove that every selected
coordinate remains dispensable throughout the generation process. An intervention can reduce
continuation locally yet still disrupt useful computation if it remains active at
later states, as demonstrated by persistent pruning failures.

The selected subspace may contain several functions. Some coordinates may support
repetition after an answer is formed, while others may contribute to verification, revision
after a conflict, or exploration of another answer. Current experiments do not
identify circuits for these proposed functions. The completion and continuation
objectives may select overlapping or state-dependent computations rather than independent
modules. A finer decomposition would require matched causal interventions across tasks
and generation states. Activation correlations or labels based only on final behavior
would not provide that evidence.

The copy-on-write probe temporarily removes the target coordinates, measures the response of the current state, and discards the intervened branch. If the rule rejects the exit,
generation continues from the untouched cache. The response measures the state's
dependence on the selected coordinates without classifying them as permanently
unnecessary.

\section{Detailed limitations}
\label{app:limits}

Terminal 8B confirmation covers one Qwen3 family, one 29.9847\%-sparse checkpoint,
MATH, two decoding seeds, and one sampling setup. Seeds 44/45 reuse the same prompts,
and the dense endpoint is historical and post-confirmation. The initial Qwen3-1.7B study
supports mechanism transfer yet it reveals failed MATH safety rules. The multiple-choice
experiments broaden the evaluation to include 13 non-math domains and two disjoint 1,024-question
populations. However, they still use the
same smaller Qwen3 checkpoint, one sampling setup, and only two decoding seeds. No new
architecture family, language, open-ended task, or powered category-specific safety
claim is tested.

Since the method copies an already completed candidate, it measures avoided reasoning-token
forwards rather than answer regeneration under the intervention. The base weight mask
does not change online, and bounded coordinate zeroing does not produce a smaller stored model. The
enumeration contract is deliberately conservative and cannot recognize every valid
answer representation. \rcc{} is not a correctness classifier: 71 of its 190 pooled
accepts are incorrect, though none convert a naturally correct trajectory into an incorrect policy outcome. The advantage of its pooled performance over post-guard accepted error remains unresolved. Zero
observed harm is finite-population evidence, not a universal guarantee. Of the 428
naturally correct 8B held-out trajectories, the one-sided exact 95\% upper bound is approximately
0.70\% under an independent-trajectory interpretation. However, shared prompts across seeds
weaken this interpretation. For the two prospective multiple-choice populations, the corresponding bounds are 0.294\% per
naturally correct source and 2.59\% among accepted naturally correct exposures.

The localization experiments measure the aggregate effect of the selected coordinates.
They do not distinguish between repetition, verification, revision, and alternative search. These possible functions cannot be inferred from activation
patterns or a small set of output examples alone. The present data also do not establish
that completion and continuation use independent subspaces.

A hardware-calibrated replay is not a production benchmark. It omits sampling,
batching, scheduling, queueing, networking, streaming, memory, energy, and sparse-kernel
effects. The full result projects directly measured primitives over all trajectories but
times only four short branch representatives end to end. Direct wall-clock latency,
batching behavior, memory traffic, and energy are not measured for the 1.7B \rcc{}
controller; its 1.145\% result is token-forward accounting only.

The experimental sequence includes adaptive development. We preserve failed tests and
keep them separate from the predefined validation and held-out evaluations. However, the
overall research program is not equivalent to one preregistered experiment.

\section{Why negative controls are part of the gate}

Matched random interventions estimate the average baseline and determine whether a state is
generally sensitive to perturbation. If the interventions substantially move the stop margin, the
target effect cannot be interpreted as mechanism-specific. The random null cap rejects
these states, even when the target minus random contrast is large. A third, disjoint
random family provides an equal-cost, sequence-level sham that does not weaken the
two-family calibration rule.

\section{Exact terminology}

A selected neuron coordinate is an intermediate MLP coordinate at the input of \texttt{down\_proj}. In earlier experiments, one directed scalar weight was edited from an
intermediate coordinate into one residual-stream coordinate. We use ``scalar MLP path''
as shorthand, not as a claim of an end-to-end semantic circuit. The final online method
is activation pruning: it temporarily zeros selected intermediate coordinates during
bounded stop probes and, for multiple choice, during a five-token forced-answer readout. This method does not apply the intervention to free-running generation.

\section{Reproducibility and resource disclosure}

\paragraph{Pinned assets.}
The primary model is \texttt{Qwen/Qwen3-8B} at commit
\texttt{b968826d9c46}; the transfer model is
\texttt{Qwen/Qwen3-1.7B} at commit
\texttt{70d244cc86cc}. Both use BF16, SDPA, and
the model's thinking mode. MATH-lighteval is pinned to commit
\texttt{0530c78699ea}; MMLU-Pro is pinned to
\texttt{b189ec765aa7}. Full 40-character hashes are retained with the run metadata.
The original Qwen3 models
use Apache-2.0 and the original MATH repository and MATH-lighteval distribution use
MIT licenses. GSM8K is cited at first use; its archived source repository does not
display a detected license, so redistribution of its examples is excluded from the
public release.

\begin{table}[ht]
\centering
\small
\setlength{\tabcolsep}{4pt}
\begin{tabularx}{\textwidth}{@{}L{2.1cm}L{2.4cm}L{2.8cm}Y@{}}
\toprule
Experiment & Model / precision & Accelerator & Purpose \\
\midrule
8B MATH evaluation & Qwen3-8B / BF16 & one A100 80GB PCIe & Validation and held-out evaluation \\
8B diagnostics & Qwen3-8B / BF16 & one L40S & Seed, latency, and endpoint analyses \\
1.7B localization & Qwen3-1.7B / BF16 & one L40S & Model-specific causal localization \\
1.7B MATH evaluation & Qwen3-1.7B / BF16 & one L4 & Controller development and confirmation \\
1.7B non-math evaluation & Qwen3-1.7B / BF16 & one L4 & Development, validation, and confirmation \\
\bottomrule
\end{tabularx}
\caption{Recorded compute environments. All runs report Python 3.11.12,
PyTorch 2.8.0+cu128, and CUDA 12.8. Exact elapsed time was not retained for every
historical run, so total project GPU-hours are not estimated.}
\label{tab:resources}
\end{table}

\paragraph{Sampling and uncertainty.}
The final 8B held-out rollouts use temperature 0.6, top-$p$ 0.95, top-$k$ 20, an 8192-token
reasoning cap, batch size 32, and decoding seeds 42 and 43. Each reported 95\% interval
uses 20,000 nonparametric bootstrap draws. Within-seed intervals resample trajectories;
the primary two-seed intervals resample prompts and keep both seed trajectories together.
We do not assume normally distributed effects. All trajectories, including wrong,
truncated, rejected, and harmed cases, remain in the estimand.
The two prospective multiple-choice populations use the same 8192-token cap and batch
size 32 under seeds 56 and 57. Their trajectory intervals also use 20,000 bootstrap
draws. The pooled analysis is descriptive because the two populations retain their
separate validation and confirmation roles.

\paragraph{Verification materials.}
The public repository provides frozen configurations, identity-clean derived records,
reference code, and scripts for auditing selected reported statistics:
\artifacturl. It contains 2,656 numerical records and supports CPU-only
recomputation of record-based controller outcomes, bootstrap intervals, and correctness
AUROC, as well as regeneration of the evidence figure from reported aggregates.
The MATH audit reapplies the frozen acceptance rule; the multiple-choice audit uses
stored controller-exit flags to reconstruct outcomes and compute accounting.
The released records do not independently reproduce every reported aggregate or
regenerate all controller decisions from model inference.
When using the multiple-choice reference helper, the caller must enforce the
prefix-level eligibility conditions, including the 64-token dwell, the 3231-token
position bound, $h_B<0$, and repeated assertions of the same candidate.
The repository does not redistribute model weights, the exact pruned checkpoint or
mask, neuron-score tensors, benchmark questions, reference answers, or model generations.
An end-to-end rerun of localization and model inference additionally requires the
omitted artifacts and original runners or their reimplementation; the pinned public
assets alone are insufficient. The software code is released under the MIT License.
Third-party models and datasets remain subject to their own terms; the accompanying
paper is licensed separately.

\section{Societal impact}

The method is designed to reduce inference costs without replacing the generator. Its
main deployment risk is silent, premature termination. Even when an individual correct trajectory is harmed, aggregate accuracy can improve, as demonstrated by the
initial 1.7B MATH diagnostic. However, a controller trained on mathematical answer syntax may fail when encountering a
new language, domain, answer interface, or model family. We do not recommend using the present rule for medical, legal, financial, safety-critical, or other high-consequence decisions. When deploying the method, it is important to preserve abstention and fallback to the
untouched route. Additionally, it is crucial to monitor natural-correct harms on application-specific data and disable
the controller under distribution shift. Any efficiency gains should not be presented as a
guarantee of correctness or as evidence that the identified coordinates form a complete
semantic circuit.

\end{document}